# SWITCHING DYNAMICS OF SINGLE AND COUPLED VO$_2$-BASED OSCILLATORS AS ELEMENTS OF NEURAL NETWORKS

A. A. Velichko[*], M. A. Belyaev, V. V. Putrolaynen, A. L. Pergament and V. Perminov

*Department of Physics and Technology, Petrozavodsk State University,*
*33 Lenina Str., 185910 Petrozavodsk, Russia*
[*]*velichko@petrsu.ru*

In the present paper, we report on the switching dynamics of both single and coupled VO$_2$-based oscillators, with resistive and capacitive coupling, and explore the capability of their application in oscillatory neural networks. Based on these results, we further select an adequate SPICE model to describe the modes of operation of coupled oscillator circuits. Physical mechanisms influencing the time of forward and reverse electrical switching, that determine the applicability limits of the proposed model, are identified. For the resistive coupling, it is shown that synchronization takes place at a certain value of the coupling resistance, though it is unstable and a synchronization failure occurs periodically. For the capacitive coupling, two synchronization modes, with weak and strong coupling, are found. The transition between these modes is accompanied by chaotic oscillations. A decrease in the width of the spectrum harmonics in the weak-coupling mode, and its increase in the strong-coupling one, is detected. The dependences of frequencies and phase differences of the coupled oscillatory circuits on the coupling capacitance are found. Examples of operation of coupled VO$_2$ oscillators as a central pattern generator are demonstrated.

## 1. Introduction

The use of artificial neural networks (ANNs)[1] in information processing has been proposed to overcome critical bottlenecks of traditional computing schemes for applications in such fields as image and speech processing, pattern recognition, and associative memory.[2] Intensive study of this problem began in the 80s of the last century with the work by Hopfield .[3] Since then, various models of ANNs have been discussed in the literature[1] for simulation of the associative memory phenomenon with learning based on the Hebb's rule.[4] Particularly, networks consisting of locally active analog elements, so-called cellular neural networks (CNNs), have been implemented in the form of integrated circuits, which are able to perform a parallel conversion of input signals for the pattern recognition problems.[5] One should also note the models of oscillatory neural networks (ONNs), e.g., the model of weakly coupled phase oscillators (Kuramoto model)[6] based on a quasi-periodical impact on inter-element connection weights, whose spatial distribution contains a certain image. In the Kuramoto model of coupled oscillators, at a certain threshold of the coupling strength, the system exhibits a phase transition: some of the oscillators spontaneously synchronize, while others remain incoherent.[6]

Upon this theoretical basis, the use of ONNs for information processing has been described in Ref. 7: the idea is to use ONNs to perform associative tasks such as pattern recognition. This is achieved by capacitive coupling of oscillators and checking their synchronization to assess the proximity of their bias. Basically, synchronization of oscillators is a crucial feature for the operation of ONNs, because the idea of synchronous oscillations[8,9] represents "a powerful instrument in designing neural networks that can perform complex information processing in agreement with known neurobiological and psychological results on the functioning of visual systems".[8]

Synchronization plays a crucial role both in elementary nervous systems (neural pairs), and in neural networks with a very large number of elements (such as e.g. cerebral cortexes).[10,11] First studies of synchronization in neural networks has been connected with the behavior analysis of the central pattern generator (CPG), which controls rhythmic movements of living organisms (breathing, locomotion, etc.). The CPG may consist of one neuron, or of a group of neurons capable to generate, even in complete isolation, certain pulse patterns. Typical oscillations produced by the CPG can be chaotic, regular, or in the form of rare bursts of spike sequences. Issues related to the neurochemical mechanisms of alternation of bursting activity, changes in the phase shift of CPG oscillations (for example, at moving of the tetrapod limbs and its modeling) remain still under debate.[10,12]

Unlike CNNs, ONNs have not yet been efficiently realized on the basis of conventional CMOS devices, and various alternative approaches are developed instead. For example, spin-torque oscillators (STOs) coupled with spin diffusion current have been proposed to construct ONNs.[13] In spite of their novelty and scalability, coupled STOs suffer from high bias currents and low speeds that are limited by the angular spin precession velocity in nanomagnets.[14] One more recently emerging trend in this area is based on synchronized charge oscillations in strongly correlated electron systems,[15] such as e. g. vanadium dioxide exhibiting an electronic Mott metal-insulator transition, which are currently considered as the most promising materials for neuromorphic oxide electronics.[16] It should be noted however that, alongside vanadium dioxide, other approaches (such as aforementioned STOs or resonate body transistor oscillators (RBO)[17]) and other correlated transition metal oxides (for instance, Ta, Ti and Nb oxides[18,19]) have also being developed and considered as prospective candidate techniques for the ONN designing.

Thin-film MOM devices based on $VO_2$ exhibit S-type switching, and in an electrical circuit containing such a switching device, relaxation oscillations are observed if the load line intersects the *I–V* curve at a unique point in the negative differential resistance (NDR) region.[20] The switching effect in vanadium dioxide is associated with the metal-insulator transition (MIT) occurring in this material at $T_t$ = 340 K.[21] Several works in the literature, including those related to the problem of ONNs, are devoted to the study of relaxation oscillations in the VO$_2$-based systems,[14,15,22-28] as well as in some other materials, phase-change memories[29] for example.

In the present paper, we report on the switching dynamics of both single and coupled VO$_2$-based oscillators, with resistive and capacitive coupling, and explore the capability of their application in ONNs. Based on these results, we further select an adequate model to describe the modes of operation of coupled oscillator circuits.

The developed model allows simulation of the operation of coupled oscillators, with a good enough approximation, in a SPICE simulator, taking into account the presence of internal noise. The numerical model fairly accurately describes the dynamic current-voltage characteristic that sets it apart from the simplified models.[25] On the other hand, physical mechanisms affecting the forward and backward switching time are identified, and this, in turn, determines the applicability limits of the proposed model. The operation of coupled VO$_2$ oscillators with strong and weak coupling, as well as an example of the circuit for simulating the generation of bursting neural activity is demonstrated.

## 2. Sample preparation and characterization

Thin films of vanadium dioxide were deposited by DC magnetron sputtering onto r-cut sapphire substrates using an AJA Orion sputtering system. The preparation procedure consisted of two stages: deposition of a vanadium oxide film at room temperature and subsequent annealing in oxygen atmosphere.

A residual vapor pressure in the vacuum chamber did not exceed $10^{-6}$ Torr, and the sputtering was carried out at a discharge power of 200 W, pressure of 5 mTorr, gas inlet speeds of 14 cm$^3$/s for Ar and 2 cm$^3$/s for O$_2$, for 20 minutes. The thickness of the obtained films was measured using an atomic force microscope (AFM) NTEGRA Prima and was ~200 nm. Annealing of the films was conducted in situ, in the magnetron vacuum chamber, at a temperature of 480°C, 10 mTorr oxygen pressure for 40-60 min.

Figure 1 shows the XRD patterns of the resulting vanadium oxide film on sapphire after annealing, obtained using a Siemens D5000 diffractometer. Before annealing, only pronounced peaks of the sapphire substrate are observed. The absence of characteristic peaks of vanadium oxides indicates that the film obtained at room temperature is amorphous. After annealing, additional peaks appear, the highest of which corresponds to VO$_2$ [011]. There are also some characteristic peaks of the highest oxide V$_2$O$_5$. Thus, the film represents a mixture of phases with the vanadium dioxide predominance.

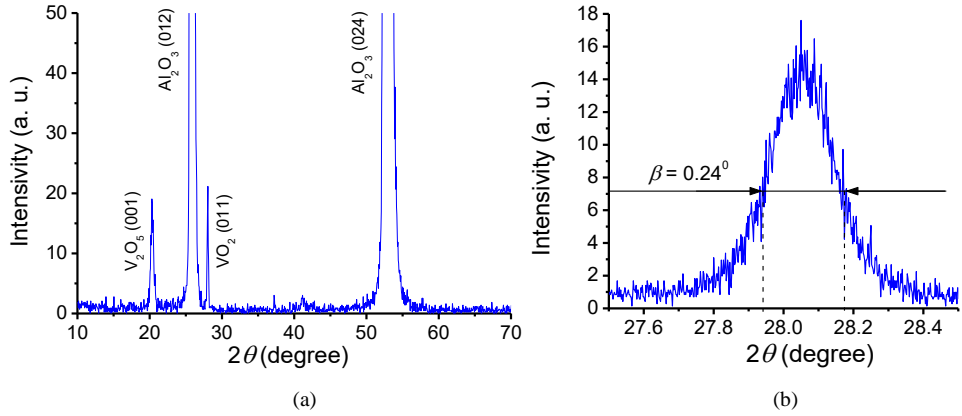

Fig. 1. XRD patterns: (a) vanadium oxide film on r-cut Al$_2$O$_3$ after annealing; (b) enlarged [011] peak of VO$_2$

A relatively low intensity of these vanadium oxides' peaks is due to the low film thickness and low degree of its crystallinity. The crystallite size *d* can be estimated from the Scherrer equation:

$$d = \frac{K \cdot \lambda}{\beta \cdot \cos(\theta)}, \qquad (1)$$

where $K = 0.9$ is the shape factor, $\lambda = 0.15418$ nm is X-ray source wavelength (Cu K$\alpha$ line), $\beta$ is line broadening at half the maximum intensity, and $\theta$ is an angle of peak maximum (Fig. 1b). According to obtained data, $\beta$ is equal to 0.24° for VO$_2$ [011] diffraction peak ($\theta \sim 14°$). Therefore calculated grain size *d* is about 40 nm.

To measure the film grain size after annealing, we have investigated the surface morphology using an AFM in a tapping mode. Figure 2a shows the surface morphology of the film obtained after annealing, in which the grain structure is clearly visible. In the grain size distribution histogram (Fig. 2b), obtained statistically in the area of 10×10 μm$^2$, one can see that the average grain size is about 200 nm. The measured average grain size exceeds considerably the crystallite size calculated from the Eq. (1). This fact can be explained by a fine structure of the grains, which might consist of several crystallites.

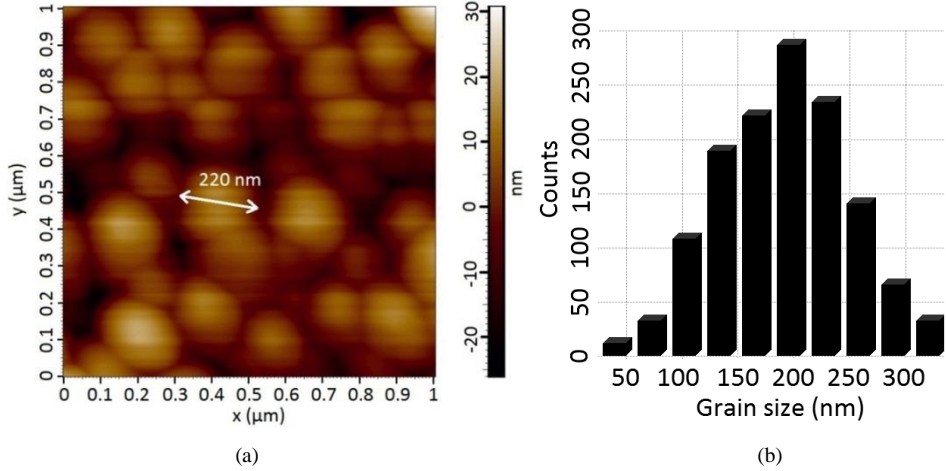

(a) (b)

Fig. 2. AFM image of the annealed film surface (a) and grain size distribution histogram

The resistivity temperature dependences of the annealed films, obtained by the four-probe method, are shown in Fig. 3. The graphs demonstrate the MIT with the resistance jump of about two orders of magnitude. Various annealing time (40-60 min) allows fabrication the films with different temperature hysteresis loops of resistance. As the annealing time increases, the proportion of $V_2O_5$ in the oxide film (whose presence is detected by XRD, Fig. 1) increases, which leads to an increase in its resistivity, transition temperature and hysteresis loop width. At a medium annealing time (~ 50 min), the transition temperature is close to that for single-crystalline vanadium dioxide, ~ 68°C. As the annealing time decreases down to 40 minutes, the film resistance becomes relatively low, the MIT is still preserved, but its temperature reduces to about 60°C. These incompletely oxidized films are of greater interest for fabrication of planar switches, since they are expected to possess a lower threshold voltage.

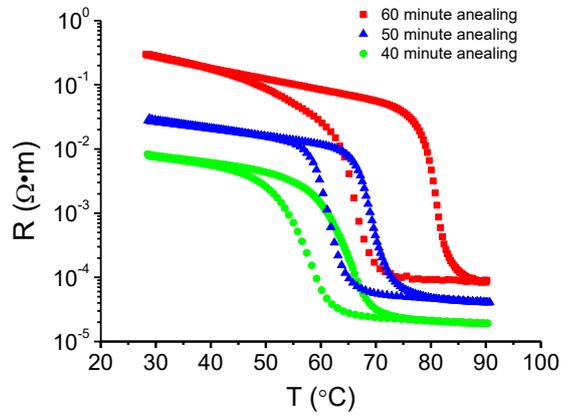

Fig. 3.  Resistivity vs. temperature for different annealing time

Planar devices were formed by the optical lithography (Heidelberg Instruments µPG 101) and lift-off processes. The electrodes were two-layer V-Au metal contacts with the overall thickness of about 50 nm (30 nm of vanadium and 20 nm of gold), and the inter-electrode gap was ~ 2.5 µm. The electrode width was ~ 10 µm. The AFM image of the obtained structure is presented in Fig. 4.

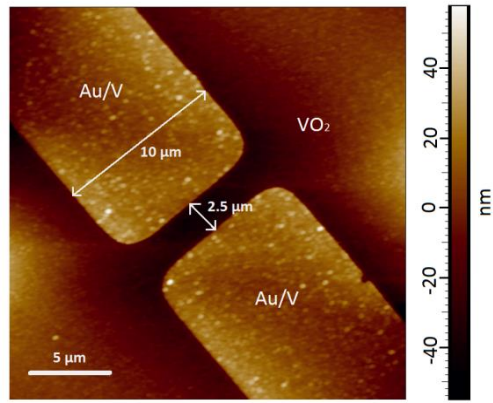

Fig. 4.  AFM image of switching structure

For electrical measurements, the samples were shielded against external electromagnetic interference, which was needed to minimize a noise affecting the circuit operation dynamics. To study the current-voltage characteristics, a sourcemeter Keythley 2410 was used. The studies of pulse characteristics and dynamics of individual (coupled) oscillators were carried out with a four-channel oscilloscope LeCroy waveRunner 44Xi-A whose maximum sampling rate was 5 GS/sec.

*Switching dynamics of single and coupled VO$_2$-based oscillators as elements of neural networks*

## 3. Experimental results

### 3.1. *I-V characteristics and transient processes of electrical switching*

A typical current-voltage characteristic of a planar switch, measured in the sweep voltage mode, is shown in Fig. 5. The current jump at the switching-on threshold voltage $V_{th}$ is ~10, and the backward jump at the switching-off holding voltage $V_h$ is ~ 5. The dynamic resistances of the ON and OFF states are ~ 40 Ω and ~ 1.1 kΩ, respectively. It should be noted that the static resistance in the ON state is nonlinear and varies from 40 to 200 Ω in the voltage range from $V_h$ to $V_{th}$.

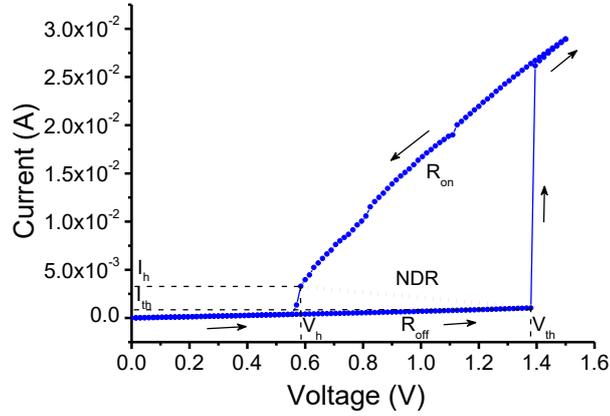

Fig. 5. Static *I-V* characteristic of VO$_2$-based switch

The parameters of this switching device are as follows: the switching-on voltage, current and power are $V_{th}$ = 1.4 V, $I_{th}$ = 1.2 mA, $P_{th}$= 1.68 mW, and those for switching-off are $V_h$ = 0.58 V, $I_h$ = 2.8 mA, $P_h$= 1.62 mW. The powers $P_{th}$ and $P_h$ are approximately equal, which refers, albeit indirectly, to the switching mechanism associated with the MIT: switching occurs at a certain critical temperature in the inter-electrode region (the model of critical temperature).[20]

The NDR region lies between the ON and OFF points (Fig. 5). Upon the films preparation, optimal parameters were chosen to provide the minimum value of $V_{th}$ and the maximum $I_h/I_{th}$ ratio. This resulted in the maximum slope of the NDR region and better stability of the ensuing oscillations. We note a high stability of the obtained switching structures, which can withstand at least over $10^9$ cycles, while preserving the threshold parameters with the accuracy of up to 5%.

The process of electrical switching is a complex physical effect involving, in addition to the threshold characteristics, temporal parameters such as the time of transition into the ON ($\tau_{on}$) and OFF ($\tau_{off}$) states. As indicated in the introduction, one of the objectives of the present work is the selection of an adequate model of the switching structure to describe the modes of operation of coupled oscillator circuits. The study of the ON and OFF times is necessary to determine the frequency range limits where one can ignore the transient processes because the switching time is relatively short as compared to a characteristic time (reciprocal frequency) of oscillations.

To study the transient processes, the delay time measurements were carried out, while a pulse with duration of 3 μs and amplitude of $V_p$ in the range of 2 to 3 V was applied to the device. Figure 6 shows an example of the transient pulse characteristics of the current and voltage variation with time. The value of $\tau_{on}$ consists of two characteristic times ($\tau_{on}=\tau_{del}+\tau_f$): the delay time $\tau_{del}$ and the pulse leading edge (front) time $\tau_f$. The time $\tau_{del}$ might vary within a wide range of 300-1000 ns, and it decreases with increasing the input pulse amplitude $V_p$, whereas the time $\tau_f$ is almost constant and equals to ~500 ns. When applying a small DC bias $V_b$ ~ 0.25 V, one can observe the switching-off delay processes occurring at the trailing edge, in the form of a gradual current decay with a characteristic time $\tau_{off}$ (Fig. 6).

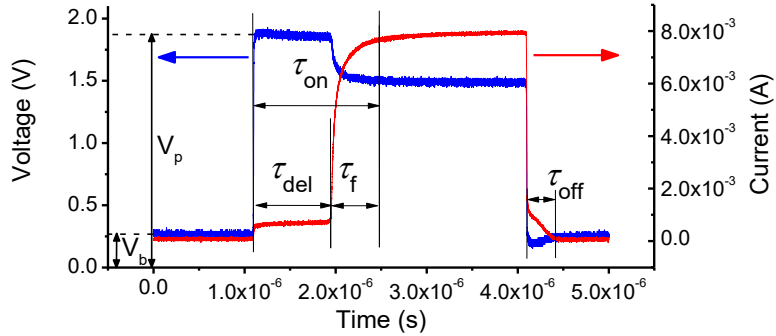

Fig. 6. Delay time measurements in the pulse mode

Figure 7 shows the graphs of $\tau_{del}$ and $\tau_{off}$ as functions of the input pulse amplitude at the 100 Hz pulse repetition frequency, as well as the dependence of the switch-off delay time on the input pulse duration at the frequency of 1 Hz.

We suppose that the delay $\tau_{del}$ is determined by the time of heating of the inter-electrode gap up to the MIT temperature in accordance with the critical temperature model.

In this case, as the input pulse amplitude increases, the power dissipated in the gap increases too, and thus the time of heating reduces. Similar results for VO$_2$-based switching structures have been obtained in the works,[28,30,31] though non-thermal electronic effects might also contribute to the switching dynamics[32-34] in high electric fields, e.g. in sandwich devices or planar devices with nano-sized inter-electron gaps. It should be noted that the time $\tau_{del}$ has a 10% relative spread, shown as the error bar in Fig. 7a, which may be associated with the threshold nature of switching and with a significant current noise. The value of $\tau_f$, responsible for the rate of rise of the current pulse, is determined by the capacitance, inductance and resistance of the switch and connecting cables; this characteristic time is constant and practically independent of the affecting pulse magnitude.

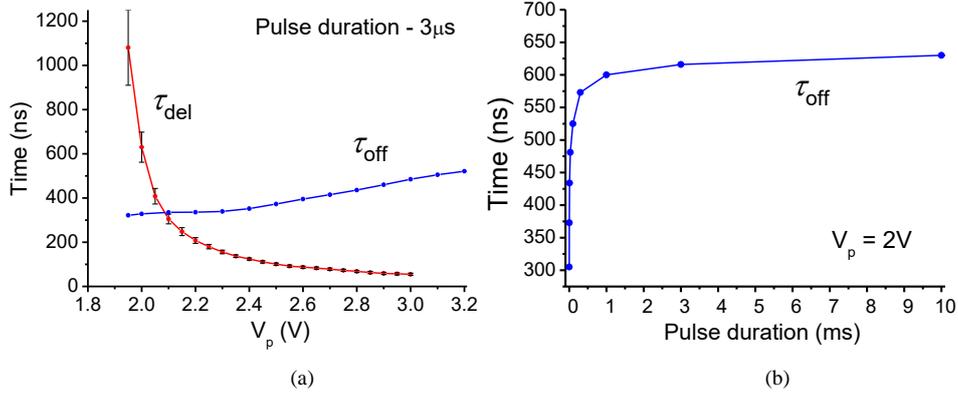

(a)  (b)

Fig. 7.  Dependences of $\tau_{del}$ and $\tau_{off}$ on pulse amplitude (a) and $\tau_{off}$ as a function of pulse duration (b)

Since there are no data in the literature concerning the switch-off time behavior, we have investigated this issue in some more detail. The value of $\tau_{off}$ is obviously determined by the time of the switching channel cooling in the metallic phase. This time linearly depends on $V_p$, at a constant pulse duration, which is due to an increase of the dissipated power and maximum temperature in the inter-electrode gap, and the higher the maximum temperature in the metal channel, the higher is the value of $\tau_{off}$.

Such a dependence of the switch-off delay time on the pulse duration (Fig. 7b) is accounted for by the fact that the maximum temperature in the channel reaches a stationary equilibrium level; as the pulse duration increases, the maximum temperature rises leading to an increase in $\tau_{off}$. When the pulse duration is higher than 1 ms, $\tau_{off}$ is almost unchanged.

Thus the characteristic times of switching $\tau_{on}$ and $\tau_{off}$, when the device is subjected to a square pulse, do not exceed the value of 1 µs. Therefore, for the oscillations with frequencies much less than 1 MHz, the times of the electrical switching transient processes can be neglected. At higher frequencies, the model of switching should take into account the foregoing experimental results.

### 3.2. *Single-oscillator circuit based on VO$_2$ switch*

To observe auto-oscillations based on a VO$_2$ switch, we have assembled a single-oscillator circuit shown in Fig. 8a. The series resistance $R_s$ and power supply voltage $V_{DD}$ are chosen so that the load line passes through the NDR region (see Fig. 5), which is a condition for the oscillations to occur, and the current resistor value $R_i$ is negligible as compared to the resistance of the whole circuit.

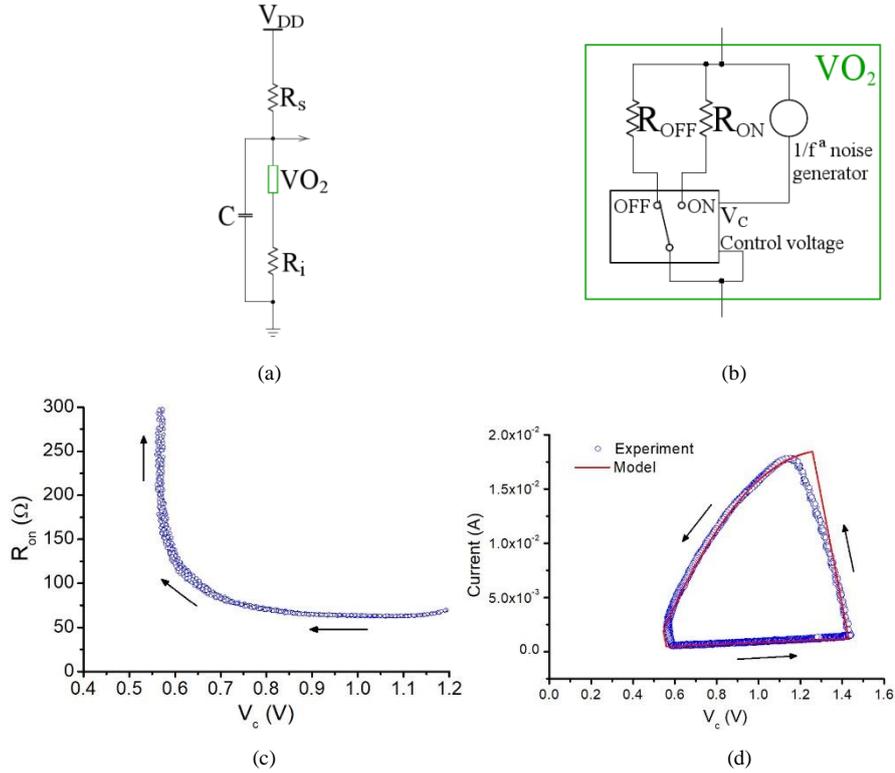

Fig. 8. (a) Equivalent circuit of single oscillator $R_i$=10 Ω, $R_s$=50 kΩ, $V_{DD}$ ~82 V. (b) Model of electrical VO$_2$-switch. (c) Resistance of the switch in the metal phase as a function of $V_c$. (d) Experimental and simulated (using LTspice software) dynamical *I-V* curves of a single oscillator

The model of a VO$_2$-switch is developed (Fig. 8b), which allows quite accurate simulation of the switching structure behavior, coinciding with the experimental data, both qualitatively and quantitatively. Switching from the ON state to the OFF state occurs when the threshold parameters of the control voltage $V_c$ are reached, and a noise generator is added to the model, simulating the pre-threshold noise observed experimentally. The maximum amplitude of the modelled noise is ~1 mV, and the spectral dependence in the range 1 to 1000 Hz corresponds to the $1/f^\alpha$ flicker noise.[35] The resistance $R_{off}$ ~ 1 kΩ is independent of the voltage $V_c$; the metal phase resistance $R_{on}$ varies with $V_c$ (Fig. 8c) and is approximated by the following expression:

$$R_{on} = \frac{V_c}{a \cdot V_c^2 + b \cdot V_c + c}, \tag{2}$$

The coefficients in Eq. (2) are fitted using the experimental current-voltage characteristics for each of the simulated switching device, and for the switch shown in Fig. 8d, they are $a = -0.031$, $b = 0.079$ and $c = -0.032$. This allows simulation of the *I-V* curve's ON branch (Fig. 8d) more accurately as compared to the work.[25] The main parameters of the tested switches differ by no more than 10%. To simulate the oscillations, LTspice software for electrical circuits has been applied. A voltage-controlled switch and programmable resistances are used (Fig. 8b) as modeling elements.

Figure 8d shows the experimental and simulated dynamic *I-V* curves of the switch obtained during oscillation at a frequency of ~7 kHz; as one can see, they practically coincide. It should be noted that the static current-voltage characteristic (Fig. 5) and the dynamic one (Fig. 8d) also differ only slightly, although these differences may increase with the oscillation frequency, which is primarily due to the substrate heating effects and the finite switching time described above. The most obvious difference between the static and dynamic experimental *I-V* curves is the linear and quadratic dependence of resistance in the metallic state. As was shown above (Fig. 7b), the cooling time $\tau_{off}$ undergoes saturation at $t$ ~ 1 ms; this time is much longer than the time of the capacitor discharge through the switch (the lifetime of the metallic phase) which is estimated to be ~ 15 μs. Therefore, the low-impedance branch of the dynamic *I-V* curve, at auto-oscillations, is in a thermodynamically non-equilibrium regime, which causes its deviation from the linear law, and it is described by Eq. (2).

Figure 9 shows typical current and voltage waveforms of the switch, as well as the oscillation spectrum. Self-oscillations resemble the Pearson-Anson effect,[36] and they appear at sequential alternation of charging of the capacitor *C* through the load resistance up to the switching-on voltage $V_{th}$ and its discharging through the resistance $R_{on}$ of the switch in the ON state. The discharging of the capacitor $C$ is accompanied by a current pulse and occurs for a characteristic time of 15 μs (for $C \sim 100$ nF). The fundamental harmonic corresponds to a frequency $F_1 \sim 7$ kHz, and the main part of the spectrum lies in the range of 0 to 30 kHz.

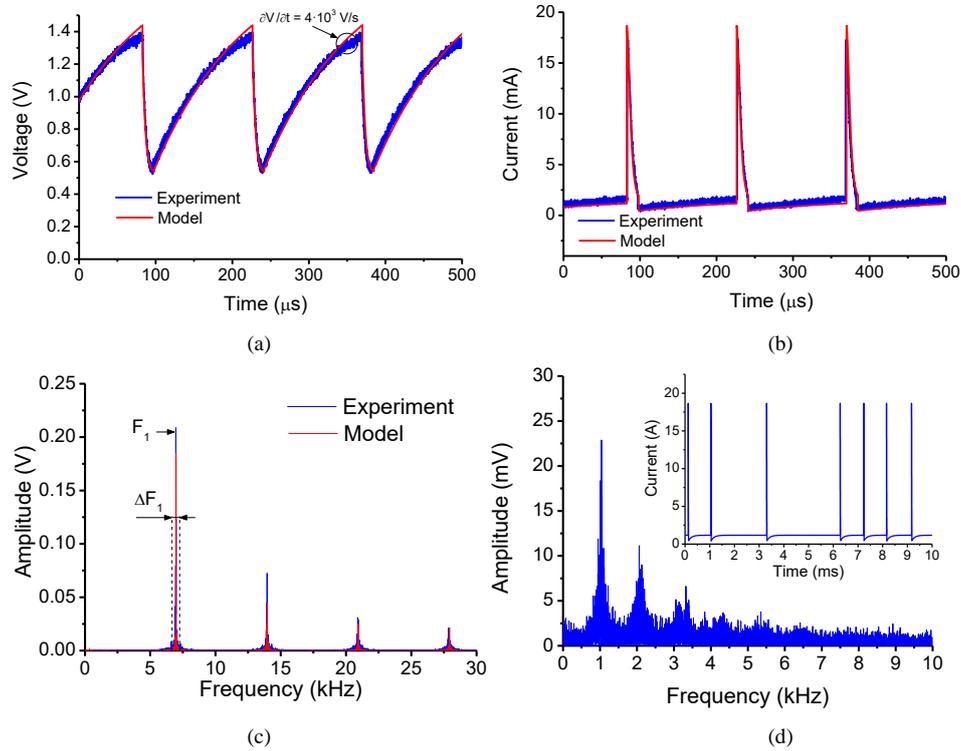

Fig. 9. Experimental and simulated relaxation oscillations of voltage (a) and current (b) for a single oscillator. Spectrum of auto-oscillations of a single circuit (c). Model signal spectrum at lowering the supply voltage down to $V_{DD} = 62$ V (d).

The dependences of the oscillation frequency on the values of series resistance $R_s$ and supply voltage $V_{DD}$ have been studied in many papers,[14,15,19] a parametric diagram has the regions of the existence of oscillations and the stay of the switch in stationary ON or OFF states. Such modes are observed in the present work too.

However, the nature of the voltage spectrum is worthy of attention (Fig. 9c): it is a discrete spectrum with the width of the 1st peak base of ~ 560 Hz (relative frequency fluctuation is $\Delta F_1/F_1 \sim 8.1 \cdot 10^{-2}$). The $\Delta F_1$ width is measured at 1% of the peak height. The broadening of the peaks is associated with the presence of noise in the subthreshold region as described above, resulting in a slight auto-oscillation frequency modulation by a chaotic signal. The spectrum of simulated oscillations also has an analogous form with the broadening $\Delta F_1 \sim 330$ Hz and relative fluctuation of frequency $\Delta F_1/F_1 \sim 4.7 \cdot 10^{-2}$, which evidences the adequacy of the proposed switching model.

If the operating point, which is in the NDR region, approaches the point of switching from the OFF state to the ON state (by means of, for example, lowering the $V_{DD}$), then the observed irregularity of oscillation is greatly enhanced. The oscillation spectrum in this case may have the form shown in Fig. 9d with the width of the peak $\Delta F_1 \sim 1.6$ kHz and relative frequency fluctuation $\Delta F_1/F_1 \sim 1.6$. The current pulses with significantly fluctuating period are observed (Fig. 9d, inset), and the dispersion of the period of such quasi-periodic oscillations has been described in our previous work.[37] This regime is also described by the proposed model, and the simulated and experimental spectra of quasi-periodic oscillations agree qualitatively.

Experimentally observed self-oscillations of the single circuit resemble irregular sequences of spikes of a pacemaker neuron, and this circuit may serve as, for example, a simple qualitative model of the respiratory rhythm.

### 3.3. *System of two R-coupled oscillators at adjacent frequencies*

The resistive coupling is rarely used for the connection of two oscillators, in contrast to the capacitive coupling playing the role of a high-pass filter. However, its advantage is obvious when scaling a circuit. That is why it has become the subject of our study. Figure 10 shows the equivalent circuit diagram of two R-coupled oscillators.

Coupling resistance consists of two resistors connected in parallel (inset in Fig. 10), viz. an external variable resistance $R_{EXT}$ and a constant one $R_{OX}$. The latter is the experimentally determined resistance between the two switches, $\sim 10$ k$\Omega$, which is due to the design features, i.e. the mutual arrangement of the switches on the continuous VO$_2$ film.

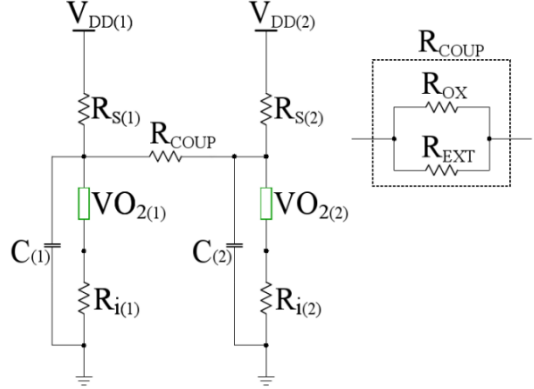

Fig. 10. Equivalent circuit of resistively coupled oscillators: supply voltages of the oscillators are $V_{DD(1)}=V_{DD(2)}=82$ V; load resistors $R_{s(1)}=R_{s(2)}=50$ kΩ; series capacitances $C_{(1)}=C_{(2)}=100$ nF; current resistors $R_{i(1)}=R_{i(2)}=10$ Ω; external resistance $R_{EXT}$ varies in the range ~ 2-10 kΩ; resistance of oxide film is $R_{OX}$ ~ 10 kΩ; resistance of coupling is $R_{COUP} = R_{EXT} || R_{OX}$ (inset).

The circuit parameters are chosen so that the oscillator frequencies differ only slightly. The natural frequencies of the individual oscillator first harmonics are $F_{1(1)}$~6610 Hz and $F_{1(2)}$~6970 Hz. The oscilloscope waveforms and spectra of each oscillator qualitatively coincide with the single oscillator spectrum shown in Figs. 9 (a-c).

At the minimum coupling strength between the oscillators, corresponding to the maximum resistance $R_{COUP} = R_{OX} = 10$ kΩ, voltage oscillations have the form similar to that of Fig.9a, and the experimental voltage spectrum of both oscillators is shown in Fig. 11. Relative frequency fluctuation for both oscillators is $\Delta F_1/F_1$ ~ $8.5 \cdot 10^{-2}$, that is close to the obtained values for the single oscillator. The inset in the Fig. 11 shows the experimental (left) and modeled (right) phase portraits. Since the oscillator frequencies do not coincide, the phase difference between the oscillators' voltages is permanently altering, whereby the phase portrait is completely filled, which is clearer observable in the simulated phase portrait of the oscillations (see right inset in Fig. 11). Thus, at the minimum coupling strength, neither frequency nor phase synchronization of the oscillator circuits is observed.

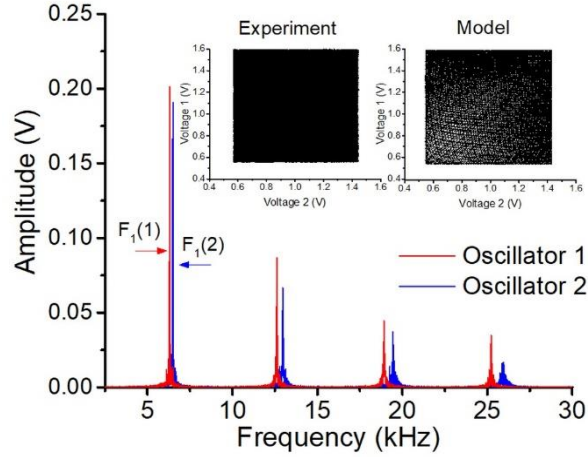

Fig. 11. Spectra of voltage signals of weakly coupled oscillators ($R_{COUP}$ = 10 kΩ). Inset – experimental and simulated phase portraits of oscillations (for building the phase trajectories, no less than 1300 oscillations were used).

When the resistive coupling is in the range 3 kΩ < $R_{COUP}$ < 10 kΩ, a gradual irregular distortion of the voltage oscillation form is observed. This change, in our opinion, is due to the bypass of one switch by another at their mutual switching to the ON state. In this range of the resistive coupling, a distortion of the spectra occurs, new harmonics emerge at mutual modulation, and the oscillations have, in general, a quasi-periodic nature (Fig.12).

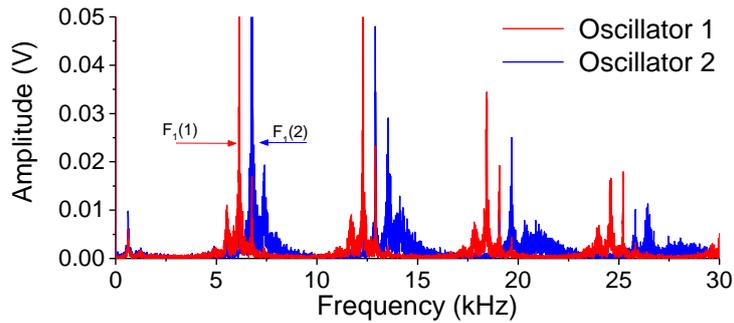

Fig. 12. Voltage spectra of resistively coupled oscillators ($R_{COUP}$ = 3.2 kΩ).

In addition, the main oscillation frequencies are reduced ($F_{1(1)}$ ~6145 Hz, $F_{1(2)}$ ~6765 Hz), which is associated with an increase in charging time of the capacitors due to leakage currents through the coupling resistor $R_{COUP}$. The experimental and modeled phase portraits reminiscent those for weakly coupled oscillators (Fig. 11, insert) without a phase synchronization.

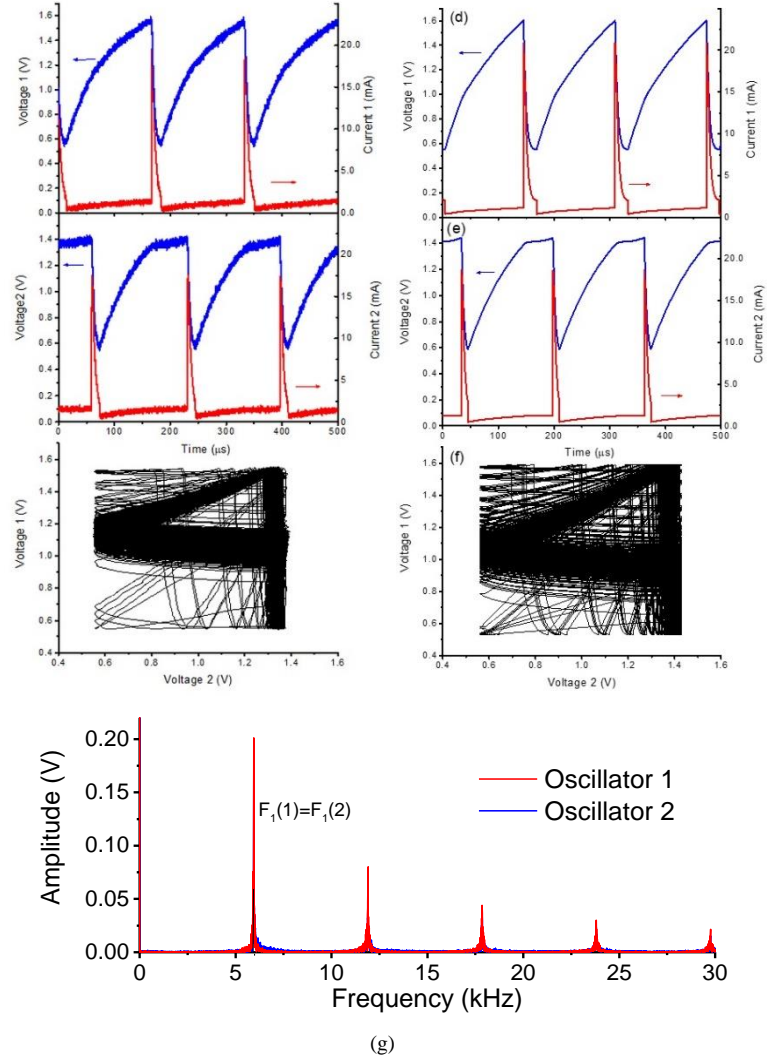

Fig. 13. Experimental (a-c) and simulated (d-f) current and voltage waveforms of resistively coupled oscillators and their phase portraits ($R_{COUP}$ = 2.4 kΩ); experimental voltage spectra of these oscillators (g).

As the coupling resistance is further decreased down to $R_{COUP}$ = 2.4 kΩ, an even greater distortion of the oscillation waveform is observed, the $F_{1(1)}$ and $F_{1(2)}$ frequencies reduce (Fig. 13), and the phase and frequency synchronization mode is set. Figure 13g shows the spectrum of synchronized oscillations, which is practically the same for both oscillators $F_{1(1)} = F_{1(2)} =$ 5945 Hz. It should be noted that the harmonics have a width of

*Switching dynamics of single and coupled VO$_2$-based oscillators as elements of neural networks*

$\Delta F_1 \sim$ 890 Hz (relative frequency fluctuation $\Delta F_1/F_1 \sim 0.15$), which is more than that for a single oscillator. On the phase portrait (Fig. 13c), one can see the main trajectory with infrequent transgressions beyond it, and we consider this effect of unstable synchronization in more detail. Figures 13d-13f presents the waveforms and phase portrait of the simulated signal. On the phase portrait, the trajectory deviations from the main path are also observed, and the voltage and current oscillation shape is close to the experimental data.

Figure 14a shows the phase shift between the coupled oscillators as a function of time. Note that for a long time interval, the phase difference is at the level of $\Delta\phi \sim 130°$ and it is punctuated by a random failure of the phase difference to a minimum. It is interesting to consider the unstable phase region (Fig. 14b): it is seen that after a phase failure, which shows itself in a premature switching of the oscillator 1 at the point A, the phase shift restores again to the original value at the point B during about ten periods of oscillations.

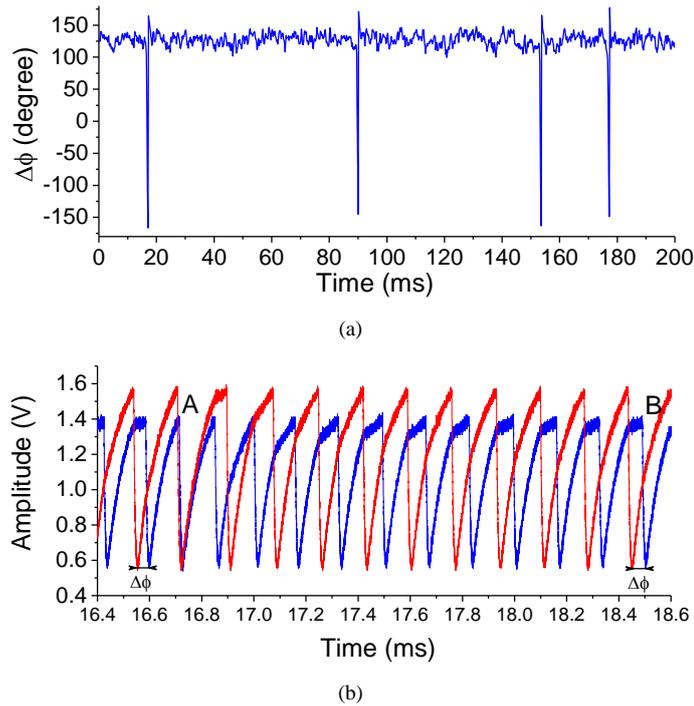

Fig. 14. (a) Phase difference of resistively couple oscillators ($R_{COUP}$ = 2.4 kΩ); (b) Region of unstable phase difference

A similar effect is observed at the numerical simulation of the circuit. The simulation shows that the phase instability effect is observed in the absence of switch noise,[23] but the observed irregularity and phase failure rate is associated exactly with the presence of noise and depends on its parameters. At the time of power-on, the oscillators also have a finite time of the transition into an in-phase mode, during ~5-10 periods.

It should be also noted that this effect of dephasing can be corrected (eliminated), both in experiment, and in simulation, by means of local tuning of the circuit parameters ($V_{DD(1)}$, $V_{DD(2)}$, $R_{s(1)}$, $R_{s(1)}$, $C_{(1)}$, $C_{(2)}$). The phase portrait in this case will have a strictly defined trajectory.

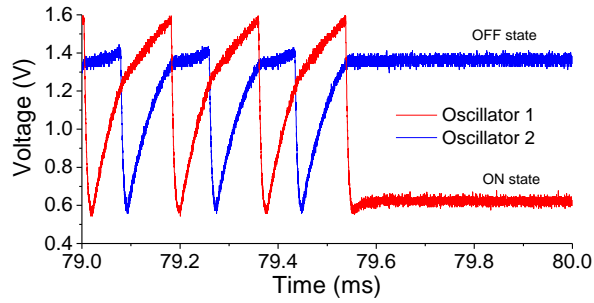

(a)

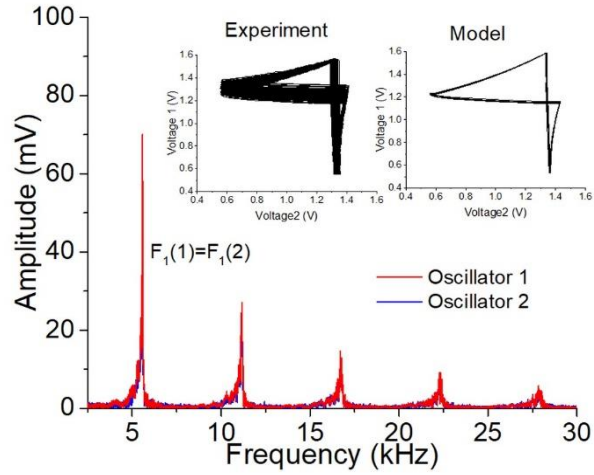

(b)

Fig. 15. Voltage waveforms of resistively coupled oscillators at $R_{COUP}$ = 2 kΩ, at the time of oscillation termination (a); experimental voltage spectrum and phase portraits of oscillators at $R_{COUP}$ = 2 kΩ (b)

As the coupling resistance is further decreased $R_{COUP} \leq 2$ kΩ, the oscillation frequency decreases, oscillations become unstable and finally disappear (after 10-100 ms of generation). Figure 15a shows the time moment of the stoppage of oscillations, after which one of the switches is in the ON state, and the other is in the OFF state. Apparently, the transition to a stable state resembles the effect of phase instability at $R_{COUP}$ = 2.4 kΩ, but no re-synchronization occurs in this case, and the operating points of the oscillators leave the NDR region to the regions of stable ON or OFF states. This effect of the oscillation suppression is observed in the simulation too. Figure 15b shows the experimental spectrum of the signal in the time interval of oscillations. The fundamental frequency is the same for both oscillators, $F_{1(1)} = F_{1(2)} = 5610$ Hz. In this mode, there is a further broadening of the first harmonic of the spectrum to values $\Delta F_1 \sim 2.1$ kHz (relative frequency fluctuation is $\Delta F_1/F_1 \sim 0.38$), which is much greater than $\Delta F_1$ for the single oscillator or coupled oscillators with a lower $R_{COUP}$.

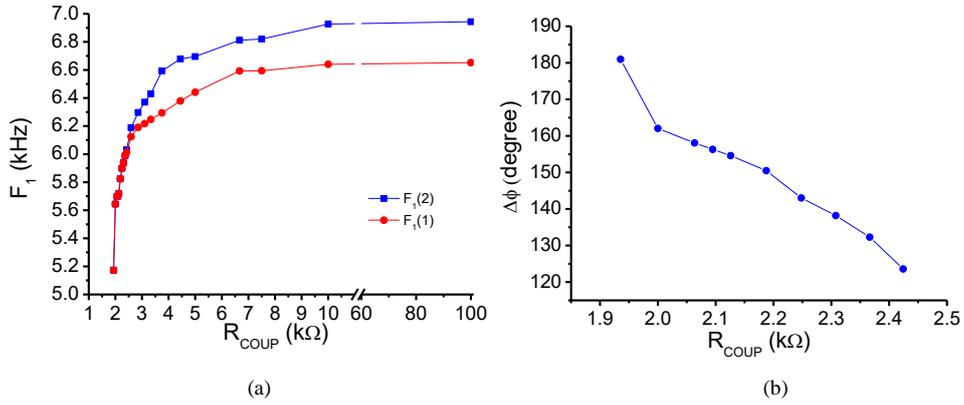

(a)    (b)

Fig. 16. Simulated dependences of frequency (a) and phase difference (b) of coupled oscillators on the value of $R_{COUP}$

Since the model is in a good agreement with experiment, the frequency and phase difference variation in a wide range of $R_{COUP}$ has been simulated. Fig. 16a represents the frequency change of the oscillator circuits as a function of the coupling resistance. As the value of $R_{COUP}$ decreases, the frequencies of both oscillators fall, and this is exactly what is observed experimentally.

When $R_{COUP}$ reaches the value of ~ 2.4 kΩ, the frequencies of oscillators are equalized, and synchronization thus occurs. At the $R_{COUP}$ values less than 1.9 kΩ, the oscillator circuits, as in experiment, are synchronized, though after several periods the switches undergo the transition into a stable state, and the oscillations cease.

The dependence of the phase difference on the coupling resistance has been investigated in the range 1.9 kΩ to 2.4 kΩ (Fig. 16b). The results show that a decrease in resistance (i.e. an increase in the coupling strength) results in a linear increase in the phase difference until the antiphase state ($\Delta\phi \sim 180°$). The state close to in-phase synchronization ($\Delta\phi \sim 0°$) is not reached.

### 3.4. *System of two C-coupled oscillators at adjacent frequencies*

Figure 17 shows the equivalent circuit of the two oscillators connected by capacitive coupling. Here we should note that the coupling is termed as capacitive rather than resistive-capacitive, although there is an embedded connecting resistance $R_{OX} \sim 10$ k$\Omega$ which is due to the design features of the mutual arrangement of the two switches. In the previous section, we found that at $R_{COUP} \geq R_{OX}$, the oscillators might be considered as non-interacting, and the presence of $R_{OX}$ resistance in the circuit has no effect on its dynamics.

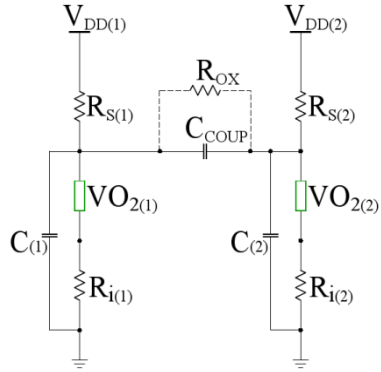

Fig. 17. Equivalent circuit of two oscillators with capacitive coupling: supply voltage – $V_{DD(1)}=V_{DD(2)}=82$ V; load resistors – $R_{s(1)}=R_{s(2)}=50$ k$\Omega$; parallel capacitances – $C_{(1)}=C_{(2)}=100$ nF; current resistors – $R_{i(1)}=R_{i(2)}=10\Omega$; $C_{COUP}$ – variable coupling capacitance (from 400 pF to 1 µF)

At $C_{COUP} < 4$ nF, the oscillators behave as uncoupled ones. Oscillation waveforms and spectra of each oscillator qualitatively coincide with those of a single oscillator (Figs. 9a-9c) with the first harmonic frequencies $F_{1(1)} \sim 6610$ Hz and $F_{1(2)} \sim 6970$ Hz.

At $C_{COUP} \geq 4$ nF, synchronization occurs. Figure 18 shows the current and voltage waveforms of oscillation and phase portraits of coupled oscillators with $C_{COUP} = 4$ nF. A qualitative agreement of the model (Figs. 18d-18f) with experiment (Figs. 18a-18c) is also observed. The spectra and phase portraits (Fig. 18g) demonstrate synchronization in phase and frequency ($F_{1(1)} = F_{1(2)} = 6090$ Hz, relative fluctuation of frequency $\Delta F_1/F_1 \sim 6.5 \cdot 10^{-2}$)

*Switching dynamics of single and coupled VO$_2$-based oscillators as elements of neural networks*

with a stable phase trajectory. The measured spectrum peaks are narrower than those for a single oscillator, in contrast to the R-coupled oscillators.

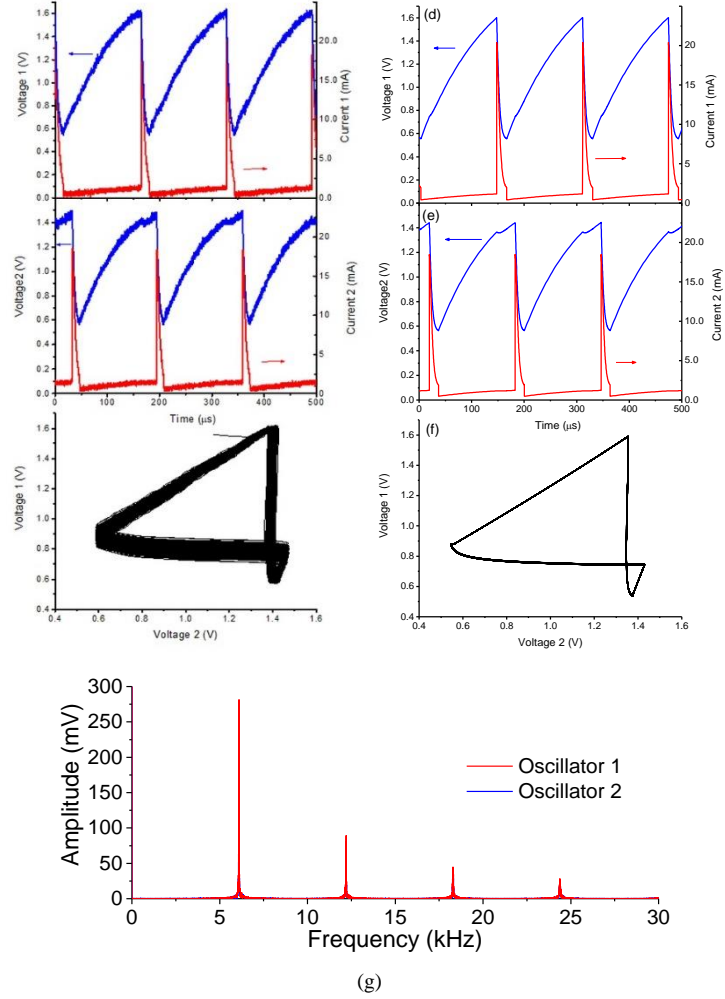

Fig. 18. Experimental (a-c) and simulated (d-e) current and voltage waveforms of C-coupled oscillators and their phase portraits ($C_{COUP}$ = 4 nF); experimental voltage spectra (g)

With increasing $C_{COUP}$, the natural frequencies ($F_{1(1)}$, $F_{1(2)}$) of oscillators are reduced, while the phase and frequency synchronization remains, but a modification of the form of oscillations occurs due to the growing mutual influence of the oscillators. As an example, Figure 19 shows the results of measurements for $C_{COUP}$=48 nF. As one can see, unlike in the case of R-coupling, the phase portraits remain unchanged at synchronization, and the effect of phase instability is not observed. The oscillators are synchronized in the state close to antiphase, which causes the appearance of double peaks in the voltage waveform and a significant reduction in the main frequencies. For example, for $C_{COUP}$=48 nF, the first harmonic frequencies are $F_{1(1)}= F_{1(2)}=$ 4655 Hz (relative fluctuation of frequency $\Delta F_1/F_1 \sim 5.9 \cdot 10^{-2}$), and the spectrum shape is similar to that at $C_{COUP} = 4$ nF (Fig. 19g).

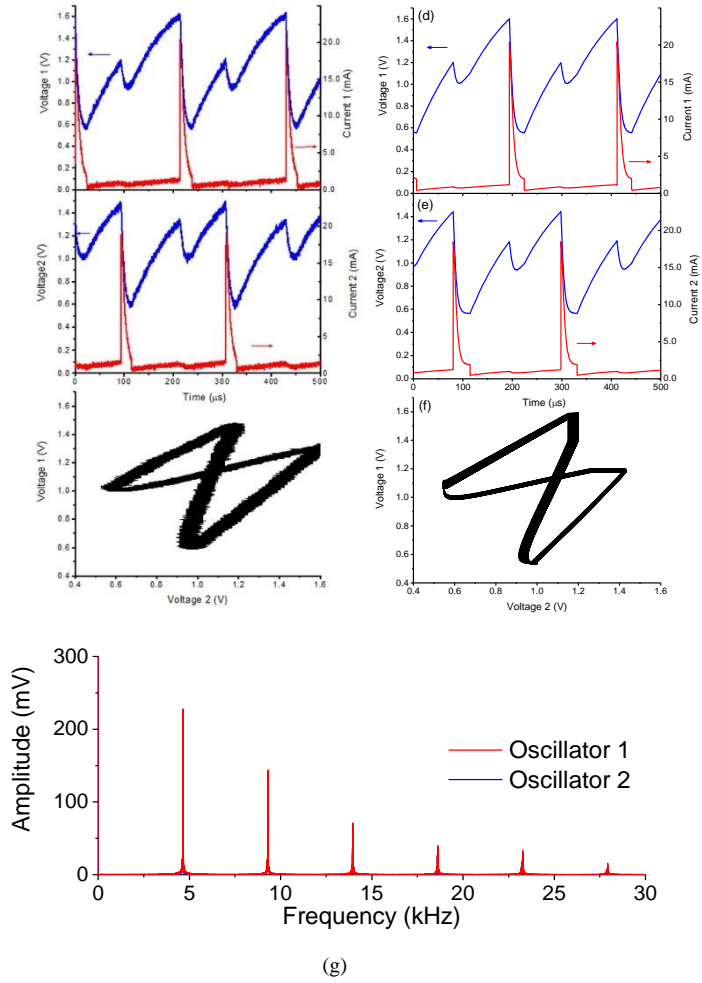

(g)

Fig. 19. Experimental (a-c) and simulated (d-f) current and voltage waveforms of C-coupled oscillators and their phase portraits ($C_{COUP} = 48$ nF); experimental voltage spectra (g)

At the coupling capacitance value 50 nF $\leq C_{COUP} \leq$ 200 nF, the oscillators get into a chaotic oscillation mode. Figure 20 shows experimental and simulated waveforms at $C_{COUP} = 90$ nF. The signal spectrum (Fig. 20g) is continuous and it is characteristic of chaotic oscillations with a relatively low average amplitude of harmonics. The phase portrait has also a complex shape, filling only a portion of the phase plane. This indicates that there are several limit cycles, between which the phase trajectory exists. We surmise that the transition to chaos occurs through the period doubling.

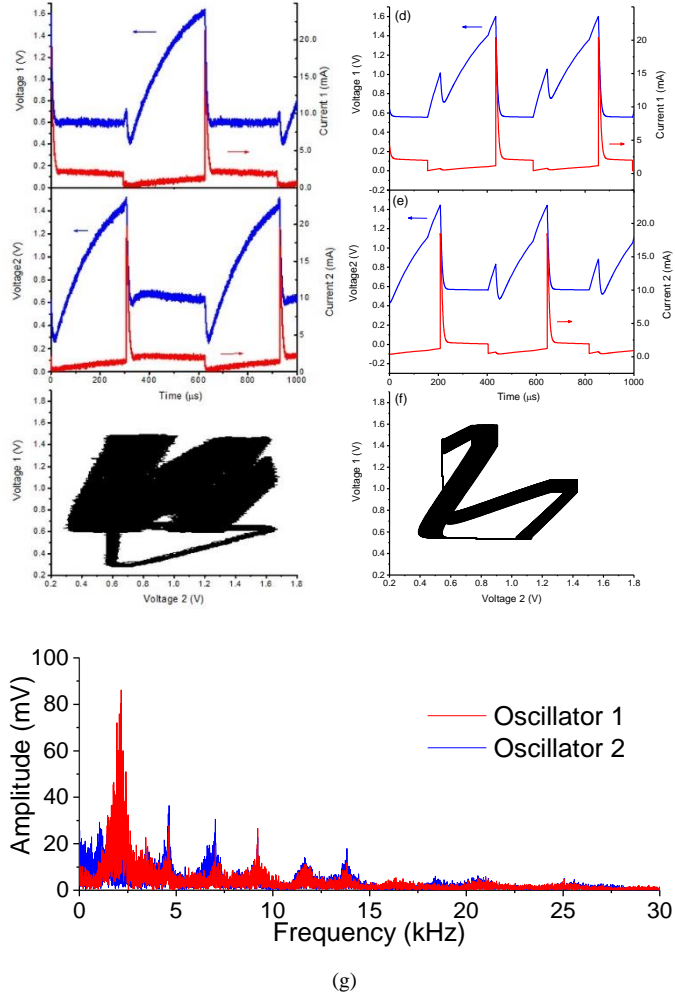

Fig. 20. Experimental (a-c) and simulated (d-e) current and voltage waveforms of C-coupled oscillators and their phase portraits ($C_{COUP}$ = 90 nF); experimental voltage spectra (g)

This mode may be termed as a chaotic transient regime between "strong" and "weak" C-coupling, because, as will be shown below, an increase in capacitance leads again to synchronization, at which the dynamics of the oscillators' operation changes considerably.

At $C_{COUP} >$ 200 nF, that is much larger than the parallel capacitances of the oscillators ($C_{(1)} = C_{(2)}$=100 nF), a transition to the synchronization mode occurs which is termed here as the "strong" C-coupling mode (Fig. 21). One of the main features of this circuit is that the decisive role in the dynamics of its operation plays exactly the coupling capacitance, and the role of the parallel capacitances is negligible.

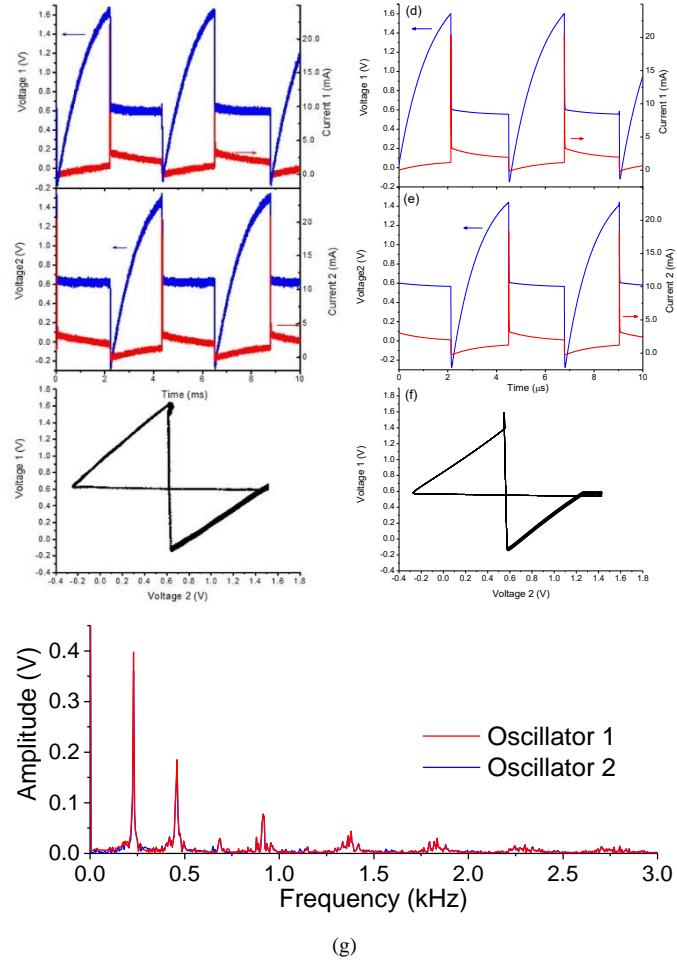

Fig. 21. Experimental (a-c) and simulated (d-f) current and voltage waveforms of C-coupled oscillators and their phase portraits ($C_{COUP} = 1\mu F$); experimental voltage spectra (g)

In this mode, the main processes at oscillations are the charging and discharging of the coupling capacitor, and it is characterized by the regions of an almost constant or slightly varying voltage (Figs. 21a and 21b). In the region of increasing voltage, as in the other modes, the switch is in the insulator state, and the coupling capacitance is charging. A longer oscillation period is due to the fact that the achieving of the switching-on voltage is determined by the charging time of the coupling capacitor, whose nominal is equal to 0.2-1µF, which exceeds the value of the parallel capacitance (0.1 µF).

When the structure is switched on, a current pulse emerges, with the duration of ~ 15 µs and amplitude ~ 20 mA, associated with the discharge of the parallel capacitance. During this same time, the coupling capacitor is also slightly discharging, but retains a large amount of charge on the electrodes. After the partial discharge of the coupling capacitor, the voltage across it reduces to ~ 0.6 V, but the switch is still in the ON state, however, its resistance increases significantly (Fig. 8c), which is caused by the features of the dynamic *I-V* characteristic. This results in a decrease of the current flowing through it almost by an order of magnitude, down to ~ 3 mA. The horizontal plateau in the obtained waveforms is associated with this small discharge current. The transition of the oscillator into the insulating state occurs when another switch is on, which promotes a decrease of the voltage on an electrode to a value less than $V_h$. Thus, this mode can not be called as "self-oscillations", since the period of one of oscillation depends entirely on the appearance of a pulse on another switch. An analogue of such a forced switching of two circuits is the operation mode of a multivibrator.

Due to the above described processes, the oscillation frequency falls to $F_1(1) = F_1(2) = 227$ Hz for both oscillators. The experimental spectrum of the oscillators (Fig. 21g) broadens as compared to that for the oscillators synchronized in the "weak coupling" mode (the relative fluctuation of frequency is $\Delta F_1/F_1 \sim 1$). The simulated waveforms of voltages and currents, as well as the phase portraits, are shown in Figs. 21d-21f. They are in qualitative and quantitative agreement with the experimental data. It should be noted that these waveforms can be modeled only provided that the Eq. (2) is taken into account, when the metal phase resistance depends nonlinearly on the voltage. The transition between the two coupling modes, within the same model, also serves as an additional proof of its relevance and applicability for further research to predict the experimental results.

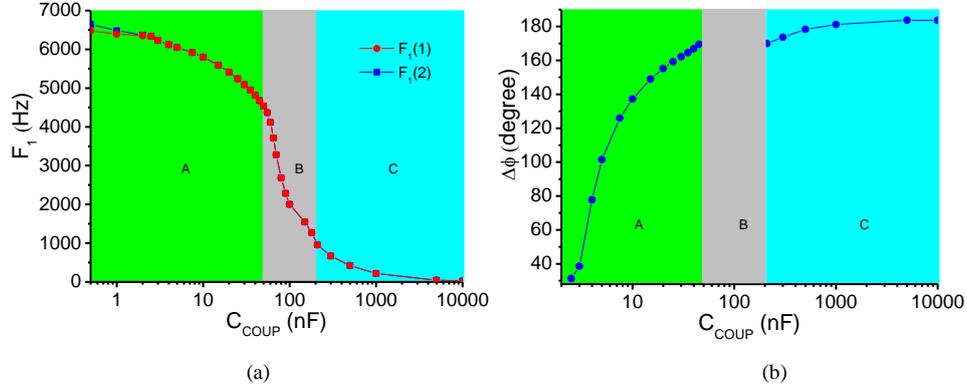

Fig. 22. Dependence of frequency (a) and phase difference (b) of the oscillator circuits on the value of coupling capacitance $C_{COUP}$: weak coupling zone A, chaotic oscillations B, and strong coupling C

Figure 22a shows the dependences of the oscillator circuit first harmonic frequencies ($F_{1(1)}$, $F_{1(2)}$) on the value of the coupling capacitance. The zones of weak coupling (A), strong coupling (C) and chaotic oscillations (B), are indicated in the plot. For the chaotic oscillation region, the average frequency of the spectrum first peak is taken. The minimum value of the coupling capacitance, at which the synchronization occurs, is $C_{COUP} \sim 2.5$ nF, and the maximum value does not have an explicit constraint (when modeling, the maximum used value of the coupling capacitance is $\sim 10$ μF). Zone (B) has a region with the maximum derivative, where the greatest influence of the coupling capacitance on the frequency is observed.

Figure 22b shows the phase difference as a function of $C_{COUP}$. In the weak coupling zone (A), as the coupling capacitance increases, the phase difference $\Delta\varphi$ raises from 30 to 170°. At the transition to the strong coupling regime (C), the oscillations are in a state close to the anti-phase one. In the regime of chaotic oscillations (B), the phase difference is unstable; therefore, its values are not shown in the plot.

## 4. Discussion

In Sec. 3.2, the dynamics of electrical switching in single structures based on $VO_2$ films has been studied. It is shown that the delay times $\tau_{on}$ and $\tau_{off}$ do not exceed 1 μs. Obviously, the switching times depend on the structure dimensions, substrate material and electrodes. For a nanoscale interelectrode gap, characteristic switching times and dissipated power should decrease significantly, and the ultimate switching speed and power are practically important. Therefore, these issues require more detailed further investigation. The presence of noise at switching is worthy of special mention. Generally, the presence of a noise in circuits is undesirable; however, it might naturally represent the properties of stochastic models of neuronal dynamics.[38] There is an assumption that the nature of the current noise in VO$_2$ switches is deterministic[35] and it leads to fluctuations of the oscillator frequency. On the other hand, a deterministic chaos in the sequences of neural spikes has been known for a long time.[10] Therefore, the presence of irregularities in the process of oscillation makes the discussed model schemes more similar to the real ONNs and CPGs.

However, the results are easier to analyze if this analysis is confirmed by a numerical model that we have implemented circuitry in the LTspice program. We have been able to show the qualitative and often quantitative agreement between experimental and simulated waveforms and their spectra, which allows the use of the model to comprehend and predict the experimental results in relatively small ensembles of VO$_2$ oscillators.

Analyzing the results on R-coupled oscillators in Sec. 3.2, we can arrive at a conclusion that the synchronization effect is observed in a relatively narrow range of resistance (Fig. 3). In this range, the dependence of oscillator phase shift and frequency on the coupling strength is almost linear. If the coupling strength exceeds a certain threshold (at $R_{COUP} < 1.9$ k$\Omega$), oscillations disappear. As is stated in Ref. 7, the phase shift dependence on the coupling parameters can be used for pattern recognition where a measure of degree of the image congruence is just the phase difference magnitude.

The narrow range of synchronization for R-coupled oscillators can be quite a positive thing, if in some application it is necessary that a circuit undergo oscillation only when the coupling resistance achieves a certain threshold. In this case, the output signal is not a phase difference or frequency, but it is mere the oscillations as such. Also, the effect of an unstable phase synchronization is characteristic of the resistive coupling at $R_{COUP} \leq 2.4$ k$\Omega$. This effect consists in a significant time-limited change in the oscillation period for one of the circuits. Thereafter, a resynchronization occurs in the same phase difference. At $R_{COUP} \leq 2$ k$\Omega$, as a result of the phase failure, the oscillation stops and the switches jump into a stable state. This phase failure seems to be associated with the synchronization instability of the coupled oscillator system itself,[25] and its aperiodicity is due to the presence of internal noise in the system.

For the resistive coupling, there is a general tendency to an increase in the first harmonic width with increasing coupling strength, and, as compared to the C-coupling, the spectra are much noisier. The maximum broadening of the spectrum corresponds to the limit of synchronization ($R_{COUP} \sim 2$ k$\Omega$). This effect, in our opinion, is primarily associated with a particular sloping shape of the voltage waveform (Fig. 13b) with a low rate of voltage rise before switching ($\partial V/\partial t \sim 6 \cdot 10^2$ V/s). Obviously, the smaller the rate of rise of voltage at the switching point, the greater the uncertainty of the time moment of switching on, and the greater an increase in the spread of the oscillation period and spectrum width. To compare, for a single oscillator (Fig. 9a), the derivative is $\partial V/\partial t \sim 4 \cdot 10^3$ V/s at switching. Other reasons for the peak broadening are, apparently, the facts that the system undergoes the unstable phase synchronization effect and that the low-frequency noises of the switches affect effectively each other through the coupling impedance, whereas, for the C-coupling, the system possesses lower volatility due to the filter property of the capacitance.

C-coupled oscillators show greater diversity of effects. Unlike for the R-coupling, for the C-coupling, the oscillations of circuits are observed in the entire range of the $C_{COUP}$ values (Fig. 22a). In addition, the three modes of coupled oscillators' operation have been found, namely the ones of weak and strong coupling, and the regime of chaotic oscillations. In the weak-coupling regime, with an increase in the coupling capacity, the frequencies of oscillators converge, and after the threshold value of $C_{COUP} \sim 2.5$ nF, the oscillators are synchronized. Unlike the resistive coupling, the phase difference after synchronization can be controllably varied over a wide range (from 30º to 170º), hence the use of such oscillators in the pattern recognition systems based on the "phase shift keying" scheme is more preferable. Also in this synchronization mode, the broadening of the spectral peaks slightly reduces in comparison with a single oscillator, and the phase difference is stable, without phase failures characteristic of the resistive coupling. In the weak-coupling mode, the effect of peak narrowing, relative to the peak widths of the uncoupled circuits, is observed; this effect of the oscillation stabilization has also been described in other studies.[13,14,39] As was already indicated for the R-coupling, the peak width strictly depends on the rate of voltage rise at the time of switching. For C-coupling, the voltage waveform undergoes such changes, accompanied by a sawtooth-like dip of voltage (Fig. 19b), that the rate of voltage rise at the time of switching eventually increases ($\partial V/\partial t \sim 6 \cdot 10^3$ V/s).

The model circuit study has revealed that at an increase of the coupling capacitance to the value of $C_{COUP} \sim C_{(1,2)}/2$ (a half the parallel oscillatory circuit capacitance, see Fig. 17), the system begins to move into the chaotic oscillation mode, and the dispersion of period and phase difference substantially increases. This allows the application of these oscillators in systems using the effect of deterministic chaos. In this case, the signal to noise ratio (SNR) can serve as a similarity metrics.[40]

With a further increase of $C_{COUP}$ up to the value of $C_{COUP} \sim 2 \cdot C_{(1,2)}$, the transition into the strong coupling mode occurs. It is characterized by forced oscillations, when the switching-on of one of the oscillators results in the switching-off of the other. In this case, the oscillation period begins to depend only on the value of $C_{COUP}$. The broadening of the oscillation peaks is much greater than that for a single oscillator, which is likely due to the prolonged staying of the operating point near the reverse transition voltage ($V_h$), where there is a significant increase in current noise.[37] On the other hand, if we use the slew rate terminology, at the OFF moment, we see abnormally low values of this rate ($\partial V/\partial t \sim 25$ V/s, Fig. 21b), which just causes a broadening of the spectra. Since this mode has a nearly constant phase difference (Fig. 22b), then its application in pattern recognition systems is questionable. However, such oscillators may be used as elements of a CPG. As was mentioned above, the simulation of the strong coupling mode is not possible without taking into account Eq. (2) describing the features of the experimental dynamic *I-V* curve. This is an important difference between the present study and other previous works on the modeling of oscillator circuits based on VO$_2$ switches.[23-26] The authors of these studies have used models with fixed values of resistances for the two phases. In our view, for a more accurate simulation of the oscillatory systems, as well as for predicting new synchronization modes, one has to take into account as much as possible features of the experimental *I-V* curves, such as noise, switching delay, nonlinear nature of the resistance, which to a large extent has been accomplished in the present study.

As a possible application of the considered systems of coupled oscillators, we propose some examples of development of a CPG. First, relaxation oscillations of a single switch circuit (Fig. 3), when admixing an external slowly varying signal to the circuit, can be transformed into a form of a burst-shaped current peak signal simulating the burst CPG mode. Preliminary analysis of the circuits shows that such a regime could be implemented with coupled oscillators too, which are significantly different in the fundamental frequencies. These modes will be the subject of our further studies.

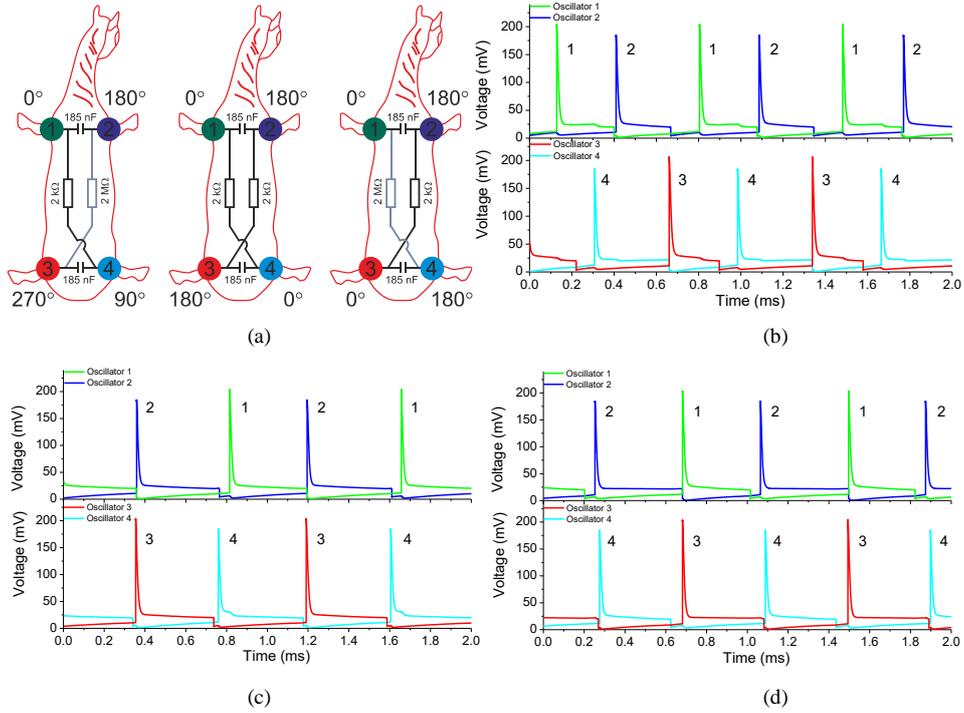

Fig. 23 Oscillators' connection scheme of CPG and phase relationships between limb movements at different gaits (a): step (I), trot (II), amble (III); voltage waveform of switches with phase shifts corresponding to the step (b), trot (c) and amble (d) gait

As another application of coupled oscillators, one can demonstrate the work of R and C couplings together in an ensemble of four oscillators to simulate changes in the phase shift of CPG oscillations when moving the limbs of tetrapods. If we connect four oscillators as shown below in the scheme of Fig.23a, then, changing only the value of coupling resistors, one can obtain different interrelations of the current oscillation phase. Also in this figure, the phase interrelations between the limb movements at different gaits are shown schematically. The simulated movement patterns are shown in Figs. 23b-23d as the waveforms of the voltage drop across the current resistors $R_i$ (Fig. 17) of the corresponding generators designated by the numbers (1, 2, 3, 4). The value of the $R_i$ resistance is small, so the voltage across it is proportional to the current flowing through the $VO_2$ switch.

Thus, the transition from one generation pattern to another can be made by changing only two coupling parameters. It should be noted that the connecting by a resistor of 2 MΩ suggests that the coupling is virtually non-existent, and these resistors are left in the circuit only for illustrative model. The generation frequency change can be achieved by varying the coupling capacitance values. In this case, an analogy is relevant with the situation when different neurotransmitter substances change the burst activity of actual CPGs.

## 5. Conclusion

In this paper, the operation dynamics of single and coupled oscillators, with both resistive and capacitive coupling, has been analyzed. Based on the data obtained, a Spice model is developed, whose adequacy is demonstrated for various operation modes of single and coupled oscillators. This model takes into account specific features of the single oscillator operation such as the threshold voltage noise and nonlinearity of dynamic I-V characteristics.

For the resistive coupling, it is shown that synchronization takes place at a certain value of $R_{COUP}$, though it is unstable and a synchronization failure occurs periodically. The dependences of frequencies and phase differences of the coupled oscillatory circuits on the value of $R_{COUP}$ are found. An increase of the width of spectral lines with decreasing $R_{COUP}$ is shown. For the capacitive coupling, two synchronization modes, with weak and strong coupling, are found. The transition between these modes is accompanied by chaotic oscillations. A decrease in the width of the spectrum harmonics in the weak-coupling mode, and its increase in the strong-coupling one, is detected. The dependences of frequencies and phase differences of the coupled oscillatory circuits on the value of $C_{COUP}$ are found. Also, the aforementioned phenomena of synchronization failure and transition to chaos via the period doubling should be considered when implementing ONNs on the basis of $VO_2$-switches.

In the present work, a CPG scheme has been simulated, which describes the different phase relations between the limbs at different types of movement of animals. We believe that the results presented in this study, concerning the dynamics of electrical switching in $VO_2$ structures, as well as various types of coupling of oscillators, will contribute to further development of artificial neural networks and their diverse applications.

**Acknowledgments**

This work was supported by Russian Science Foundation, grant no. 16-19-00135.